\documentclass{article}

\usepackage[preprint]{neurips_2021}

\usepackage{subcaption}
\usepackage{graphicx}
\usepackage[utf8]{inputenc} 
\usepackage[T1]{fontenc}    
\usepackage{hyperref}       
\usepackage{url}            
\usepackage{booktabs}       
\usepackage{amsfonts}       
\usepackage{nicefrac}       
\usepackage{microtype}      
\usepackage{xcolor}         

\title{MA-Dreamer: Coordination and communication through shared imagination}

\author{%
  Kenzo Lobos-Tsunekawa \\
  The University of Tokyo\\
  \texttt{kenzolobos@gmail.com} \\
  \And
   Akshay Srinivasan \\
   Sony AI \\
   \texttt{akshay.srinivasan@sony.com} \\
   \AND
   Michael Spranger \\
   Sony CSL \\
   \texttt{michael.spranger@sony.com} \\
}
\hypersetup{draft}
\begin{document}

\maketitle

\begin{abstract}
Multi-agent RL is rendered difficult due to the non-stationary nature of environment perceived by individual agents. Theoretically sound methods using the REINFORCE estimator are impeded by its high-variance, whereas value-function based methods are affected by issues stemming from their ad-hoc handling of situations like inter-agent communication. Methods like MADDPG are further constrained due to their requirement of centralized critics etc. In order to address these issues, we present MA-Dreamer, a model-based method that uses both agent-centric and global differentiable models of the environment in order to train decentralized agents' policies and critics using model-rollouts a.k.a `imagination'. Since only the model-training is done off-policy, inter-agent communication/coordination and `language emergence' can be handled in a straight-forward manner.
We compare the performance of MA-Dreamer with other methods on two soccer-based games. Our experiments show that in long-term speaker-listener tasks and in cooperative games with strong partial-observability, MA-Dreamer finds a solution that makes effective use of coordination, whereas competing methods obtain marginal scores and fail outright, respectively. By effectively achieving coordination and communication under more relaxed and general conditions, out method opens the door to the study of more complex problems and population-based training.
\end{abstract}

\section{Introduction}

Reinforcement Learning (RL) is concerned with the solution to Markov Decision Processes (MDP). RL, due to its end-to-end nature and 0th-order oracle requirements, has been used to solve many challenging problems in game AI (\cite{dqn,muzero}), NLP (\cite{nlprl}), and robotics (\cite{ddpg}). However, many of these problems correspond to the single-agent domain where the environment being stationary, is unaffected by the learning process. Often, however, an agent competes or cooperates with others, while also interacting with the common environment. Multi-Agent RL (MARL) deals with such problems and allows us to tackle more realistic decision processes. MARL has been applied to social problems like negotiation (\cite{negotiation18}), coordination (\cite{coordinationsoccer,otherplay}), and the emergence of language (\cite{Mordatch2018}).


In the multi-agent setting, the overall environment ceases to be stationary from the viewpoint of individual agents, since all the agents are concurrently updating their policies. For this reason, RL algorithms designed/tested for single-agent domains with stationary environments fail in the MA setting. Value-learning methods like Q-Learning (\cite{dqn}) largely ignore the non-stationarity of the process resulting in unstable learning save for the most simple problems (\cite{communication-qlearning}). On the other hand, policy-gradient (PG) methods, while theoretically capable of handling non-stationary environments, have only been used to study interesting yet simplified problems (\cite{guess-who, continuous-comms}). PG methods remain affected by the large-variance of the gradient-estimator in high-dimensional spaces (\cite{referential-image-reinforce,referential-pixel}); additionally, these methods cannot reuse past-experience of the agents, making them woefully sample-inefficient and impractical for complex tasks with high-dimensional (e.g image, lidar) observations.

In this context, \cite{maddpg} proposes Multi-agent Deep Deterministic Policy Gradient (MADDPG), a practical algorithm to solve MARL in the form of an off-policy actor-critic algorithm, that has been proven able to solve several MA tasks. However, its characteristics impose several restrictions to the problems it can solve: first, by requiring centralized critics and each agent to have access to other agents' policies, the agents' structure must remain constant during training (e.g., cannot change agents) and second, by using old and outdated transitions the performance gets reduced when using communication channels or partial observations.

In our work, we combine recent advances in Model-based RL (MBRL) to address the current limitations of MARL. We base our method on Dreamer (\cite{dreamer2}), a novel MBRL that can tackle high-dimensional and complex tasks through model-rollouts dubbed 'imagination'. 

Our algorithm learns multiple world models and extends the extend the imagination procedure to shared-imagination, which decentralizes RL learning, re-utilizing previous data to generate imaginary trajectories for policy learning, thus avoiding restrictions from previous algorithms in terms of problem structure, outdated transitions, and sample efficiency. These characteristics are particularly convenient to learn communication policies, as they allow on-policy learning with large batches.

We evaluate our method in soccer-inspired tasks, which have long been used for benchmarking different aspects of MA, such as communications and coordination(\cite{hfo, coordinationsoccer}). In particular, we address two tasks in the soccer environment: A speaker-listener task resembling classic referential games, where one agent can observe the entire field and another must play soccer having reduced information; and a standard soccer game, where agents have symmetric limited information and must act cooperatively to win the game. With these tasks, we show the effectiveness of shared imagination and its effects on communication-enabled agents and compare it with state-of-the-art MA-oriented algorithms. 

In summary, our main contributions are: A MBRL approach for MARL problems without restrictions on the agents and observation spaces with improved sample efficiency due to experience-replay support and shared imagination, a methodology that enables sequential communication learning and also acts as a task-aware coordination method outside interactions with the environment.








\section{Related Work}

The solution to multi-agent tasks/games often gives rise to interesting behavior like co-ordination and communication, and in this context, RL becomes a powerful tool to computationally study such problems and the emergence of such behavior. Both vanilla classic value-based methods and policy-gradient based methods have been used to study referential \cite{referential-image-reinforce,referential-pixel,communication-multimodal-multistep-referential} and coordination games \cite{negotiation18,Mordatch2018}, but due to either ignoring non-stationarity or large-variance estimators (larger than the rate the MDPs change), problems are usually reduced to having short time horizons \cite{guess-who, negotiation18}, reduced number of agents, full (physical) observability (\cite{Mordatch2018}), or reduced-search spaces \cite{communication-qlearning, negotiation18}.

MA-focused RL algorithms usually extend single-agent algorithms, attempting to replicate their properties while addressing the difficulties of MA. In particular, MADDPG (\cite{maddpg}) extends the DDPG algorithm (\cite{ddpg}), enabling the use of experience replays to improve data efficiency in MA. 

However, MADDPG poses several disadvantages such as policy updates using the gradient from critics trained on transitions coming from outdated MDPs, the requirements of centralized critics, and the explicit requirement of agents to have knowledge of the other agent's policies.

In our work, we address some of the limitations of MADDPG by replacing the centralization of the critics and the assumptions of known policies to allow more flexible schemes where other agents can belong to a heterogeneous population and use policy updates resembling policy gradients and rollouts rather than relying on the deterministic policy gradient and TD-learning for policies due to their beneficial properties in MARL.




Model Based Reinforcement Learning (MBRL) addresses the sample-(in)efficiency encountered in their model-free counterparts by learning a model of the underlying MDP first, instead of learning exclusively a policy - a highly-desired property in MARL due to the large spaces induced by the multiple agents. However, naive MBRL implementation for MA tasks can suffer from the perceived non-stationarity of the processes when models are coupled with policy learning, so we base our proposed method on recent advances in MBRL (\cite{dreamer2}) and propose a pipeline to benefit from the advantages of MBRL that addresses the needs of MARL.

\section{Preliminaries}
In the MA problem setting, the overall $\textrm{MDP}_{\textrm{global}}$ can be formulated with the tuple $(S, O_{1}\times ... \times O_{n}, A_{1}\times ... \times A_{n}, C_{1}\times ... \times C_{n}, R_{1}\times ... \times R_{n}, \gamma)$, where $S$ is the global state space of the environment and $\textrm{MDP}_{\textrm{i}}(S, O_i, A_i, C_i, R_i)$ are the marginal MDPs perceived by the i-th agent, composed by the corresponding i-th agent's observation ($O_i$), action ($A_i$), communication ($C_i$), and reward spaces ($R_i$), and $\gamma$ is the discount factor of the problem.

In the most general case, agents do not have access to each other actions, and the effects of the behavior of one agent can only be perceived by other agents if such actions modify the global state in a way that the marginal observation also gets affected. Since this effect may not be immediate nor apparent, MARL becomes a difficult task, since as policies evolve, the perceived marginal MDPs become non-stationary. 

To facilitate learning and enable more complex and coordinated behaviors, it is also possible to consider a free-talk communication channel of discrete symbols (\cite{referential-image-reinforce}), which can also be modeled as actions, with the characteristics that they do not modify the state of the environment but are available to other agents as part of their observations. However, the intended meaning of communication changes as the policies evolve, making the use of outdated communication codes unfit for learning, similar to the observations' case.


\section{Multi Agent Dreamer}
\label{sec:madreamer}

\subsection{World Models for MARL}
\label{sec:ma-wm}

\begin{figure}
  \centering
  \includegraphics[scale=0.55]{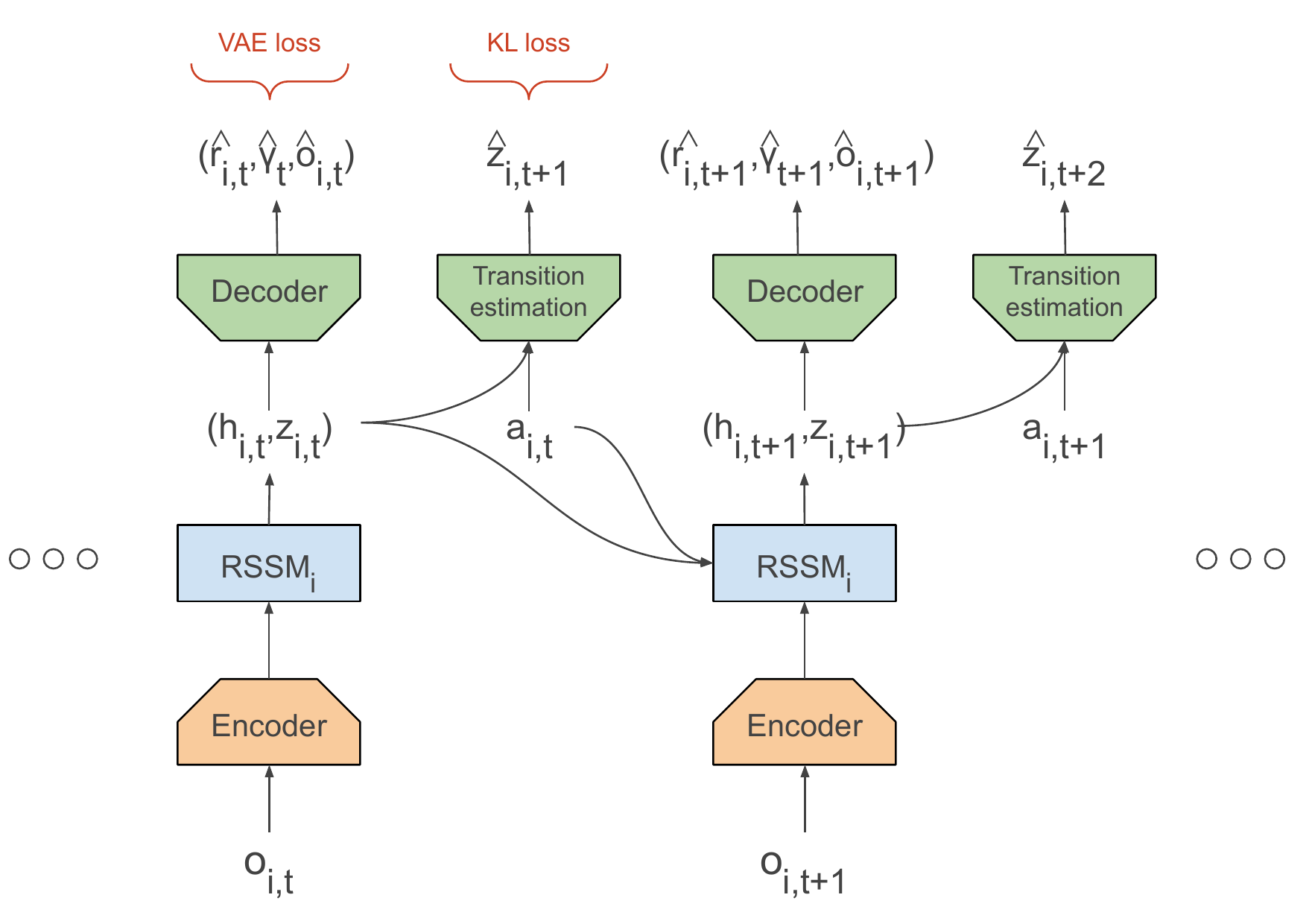}
  \caption{Agent world model.}
  \label{fig:dreamer-wm-agent}
\end{figure}

\cite{dreamer1} and \cite{dreamer2} introduces Dreamer, an MBRL method that can handle stochastic environments with high-dimensional observations. At the core of this algorithm is the world model implemented by a Recurrent State Space Model (RSSM), first proposed by \cite{planet}, which estimates the transition and observation distributions based on a latent-space model. The latent-space model is trained to reconstruct the observations, rewards, and the latent-code for the next state, given the current action, as shown in Figure \ref{fig:dreamer-wm-agent}.

\begin{figure}
  \centering
  \includegraphics[scale=0.55]{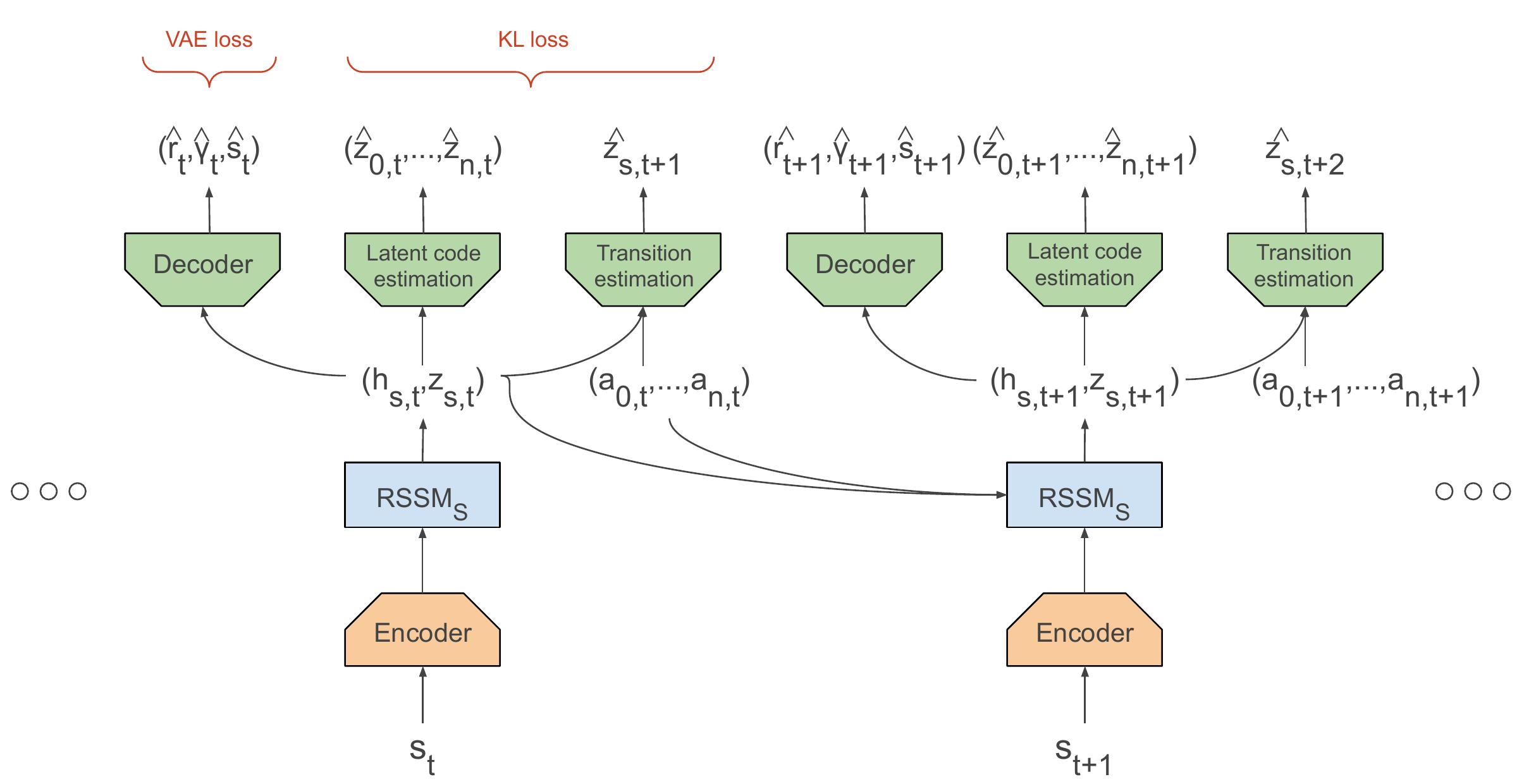}
  \caption{Global world model.}
  \label{fig:dreamer-wm-global}
\end{figure}

For the MA case, we create a standard world model per agent but do not include the communication symbols as part of the model, since they do not affect the physical environment, and an additional world model dubbed the global world model. The global model, presented in Figure \ref{fig:dreamer-wm-global} instead of using a particular observation, uses either the complete collection of observations as inputs and targets to reconstruct, or the state when available. In either case, in addition to the reconstruction of the observations, reward, and transition, the latent space is tasked to estimate the latent space of the individual agents for the same time-step, for the reasons explained in Section \ref{sec:sharedimag}.

The use of global information, although seemingly impractical, has several ways of being implemented. For example in card games, although the other players' hands are not known during play, for learning purposes, they can be accessed. Even in the case of physical environments, like sport games, the use of a bird-eye camera would represent this global information. It is also important to remark that even in the case we use global state information during training, the global model is not used during policy evaluation, enabling the use of agents based on just their local information. 

In the single-agent case, learning the model does not present problems as the MDP is stationary, even when not fully observable. On the other hand, in the case of MA, while the global world model is still learned over a stationary MDP, the individual agent's world models MDPs are not stationary due to the non-observable effect of the evolution of the other agents. While stationarity is critical for RL, it is not a limitation for world models, since they are learned through more stable supervised learning.

\section{Coordination and communication via shared imagination}
\label{sec:sharedimag}

\begin{figure}
     \centering
     \begin{subfigure}[b]{0.33\textwidth}
         \centering
         \includegraphics[width=\textwidth]{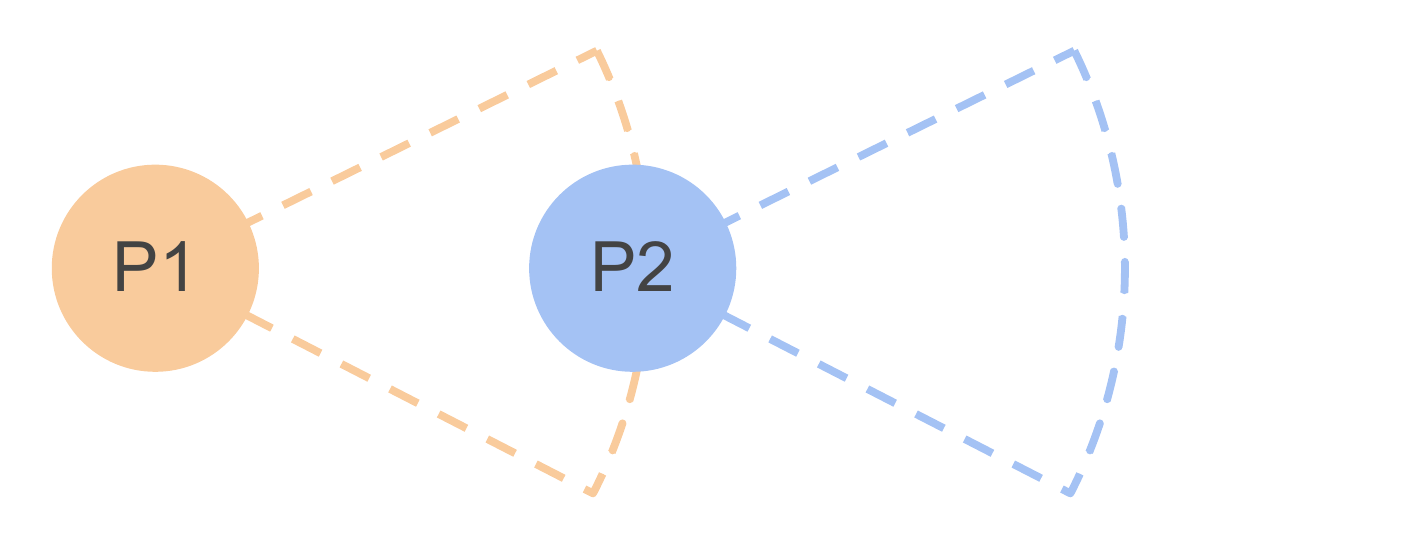}
         \caption{Initial situation}
         \label{fig:pitfall-base}
     \end{subfigure}
     \\
     \begin{subfigure}[b]{0.33\textwidth}
         \centering
         \includegraphics[width=\textwidth]{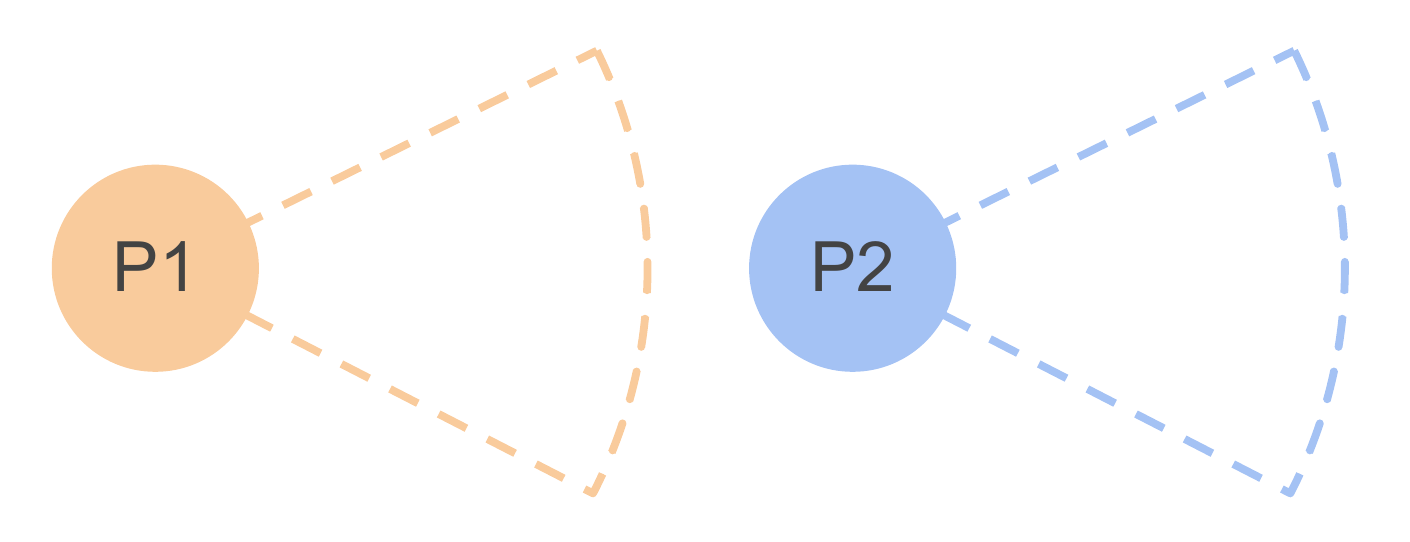}
         \caption{Player 1 unrolls its world model.}
         \label{fig:pitfall-p1}
     \end{subfigure}
     \qquad \qquad
     \begin{subfigure}[b]{0.33\textwidth}
         \centering
         \includegraphics[width=\textwidth]{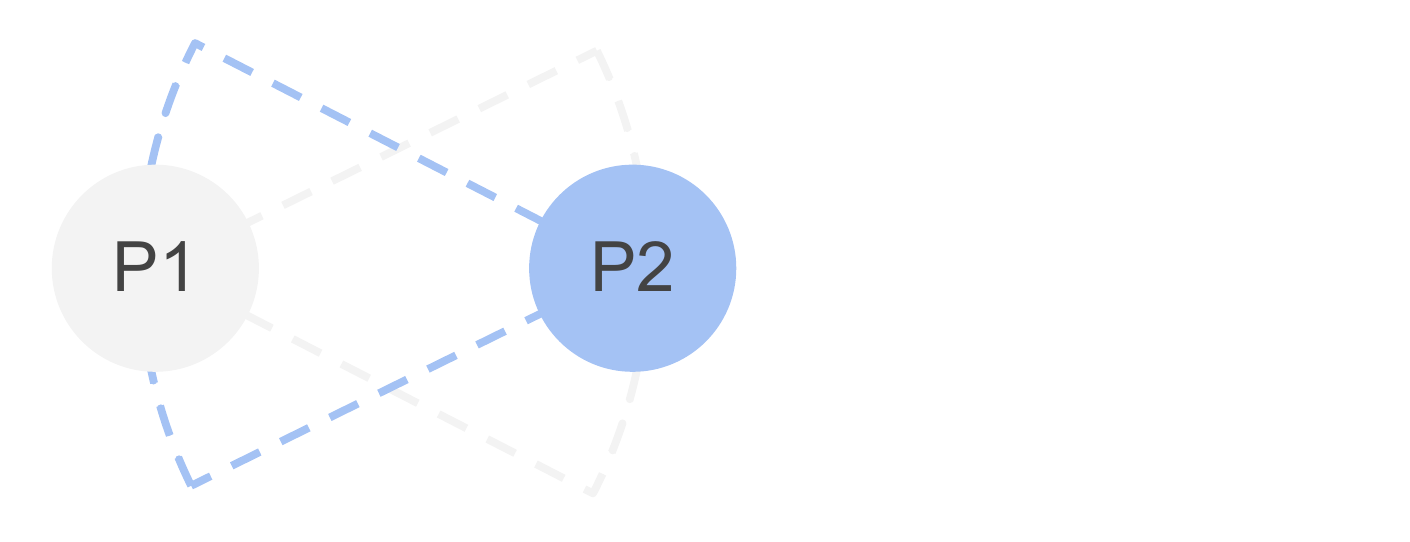}
         \caption{Player 2 unrolls its world model}
         \label{fig:pitfall-p2}
     \end{subfigure}
        \caption{Pitfalls of the naive imagination method in MA tasks. Agents are represented by circles with dashed areas representing their visual ranges in a partially observable environment. In (a), Player 1 (orange) detects Player 1 (blue) in its field of view, but not the other way around. Player 1 stays in the same position, whereas Player 2 rotates in 180 degrees. In (b), since Player 1 does not have access to Player 2's policy, it estimates that it will move forward, culminating in it getting out of its field of view. On the other hand, in (c), by unrolling Player 2's model, the agent rotates but does not detect Player 1 since that information was not available to Player 2's world model, culminating in both cases in diverging imagined rollouts.}
        \label{fig:imagination-pitfall}
\end{figure}

Imagination-based rollouts are advantageous, as they enable online policy learning with experience replays and planning in compact and information-rich latent spaces. However, using multiple single-agent versions of Dreamer in the MA setting leads to similar problems as model-free methods: independent rollouts have bad sample efficiency and using the different rollouts jointly is not possible since the imaginary trajectories diverge due to partial observability and the lack of knowledge of other agents' policies (an example of the this phenomenon is presented in Figure \ref{fig:imagination-pitfall}).

\begin{figure}
  \centering
  \includegraphics[scale=0.4]{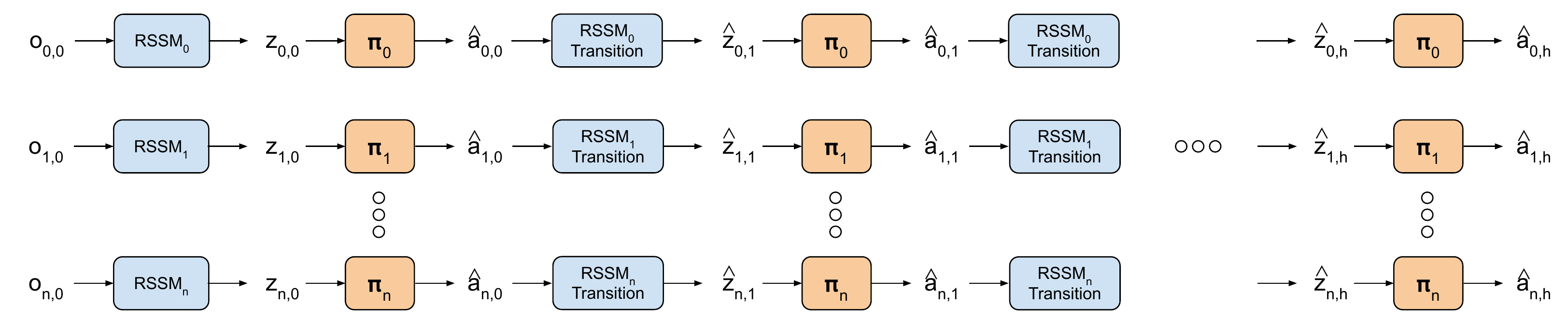}
  \caption{Naive MA imagination pipeline. Only the first observations $(o_{0,0}, o_{1,0},..,o_{n,0})$ and their corresponding latent vectors $(z_{0,0}, z_{1,0},..,z_{n,0})$ come from the real MDP, whereas all subsequent elements are unrolled in the imagined MDP. Blue and orange blocks are differentiable-learnable components, but only the policies (orange) are learned when unrolling trajectories. The process considers $n$ agents and an unrolling horizon $h$, with the rewards being are provided by the individual space latent vectors $(z_{0,t}, z_{1,t},..,z_{n,t})$.}
  \label{fig:imagination-naive}
\end{figure}

Naive imagination procedure is the result of unrolling imaginary trajectories using the marginal RSSM models illustrated in Figure \ref{fig:imagination-naive}, where each agent unrolls its model oblivious of the other agents. The main component of our proposed method 'shared imagination', aims to solve the problems of the naive approach, by providing coordination and consistency to the unrolled imaginary trajectories.

The process of shared imagination starts with the sampling of a state ($s_0$) from the experience replay. We then use the global RSSM to estimate $z_{\textrm{global},0}$, and with this information, we reconstruct the collection of latent-space vectors for all individual agents $(z_{\textrm{agent}=0,0}, z_{\textrm{agent}=n,0})$. With the individual latent vectors we apply the decentralized individual policies obtaining the collection of imaginary actions $(a_{\textrm{agent}=0,0}, a_{\textrm{agent}=n,0})$, but instead of using these imaginary actions to unroll the individual agent RSSM's, we unroll the global one, obtaining the next $z_{\textrm{global},1}$. This process is then repeated until a desired trajectory length, upon which we perform policy optimization.

The reasoning behind unrolling the global model instead of the local ones is to produce consistent latent vectors for all agents while avoiding the use of the individual agents' observations. Although unrolling the global model seems to make the individual RSSM irrelevant, since the global model estimates the individual latent spaces which were trained independently, from the perspective of the agents, the latent spaces come from their own model, making it possible to afterward deploy the individual agents without the global RSSM in a decentralized fashion. 

In this way, the global RSSM acts as the centralized point of the algorithm (akin to the centralized Q-functions in MADDPG) but is decoupled from the RL training, while allowing for the effective coordination of the agents. An additional point of note is that since at any point in time, imaginary trajectories are sampled with the latest RSSM, it allows for online policy training that never uses outdated transitions, easing  learning in general and in particular communication learning. 

\begin{figure}
  \centering
  \includegraphics[scale=0.4]{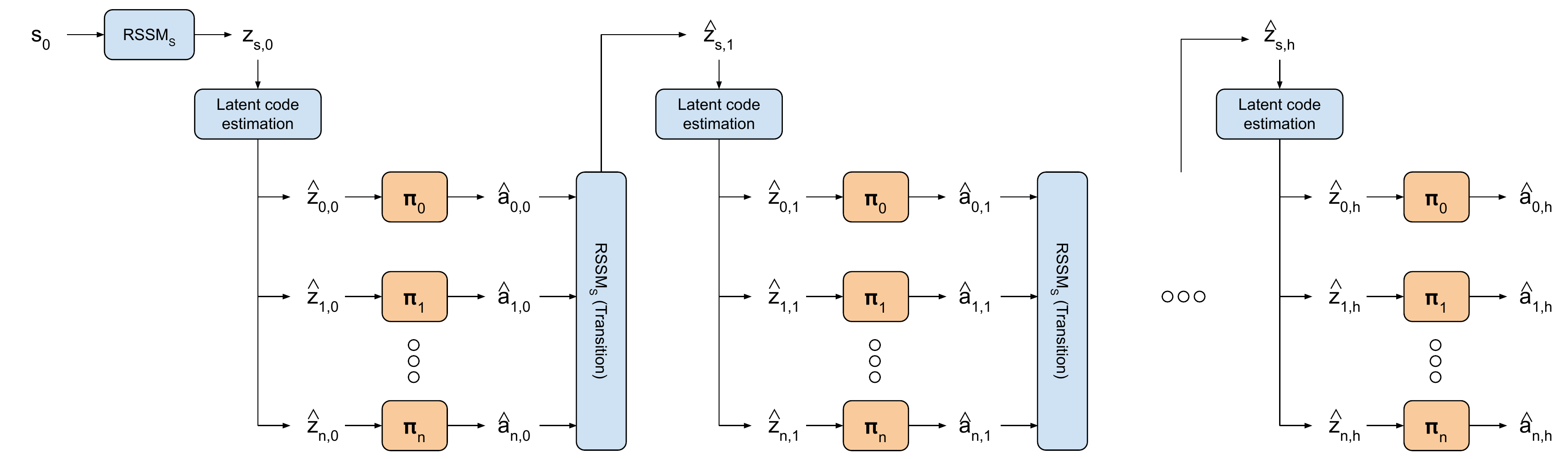}
  \caption{Shared imagination pipeline. Only the first state ($s_{0}$) and state latent vector ($z_{s,0}$) come from the real MDP, whereas all subsequent actions and latent vectors are unrolled in the shared imagined MDP. Blue and orange blocks are differentiable-learnable components, but only the policies (orange) are learned when unrolling trajectories. The process considers $n$ agents and an unrolling horizon $h$, with rewards being provided by the space latent vector $z_{s,t}$.}
  \label{fig:imagination-shared}
\end{figure}

\begin{figure}
  \centering
  \includegraphics[scale=0.5]{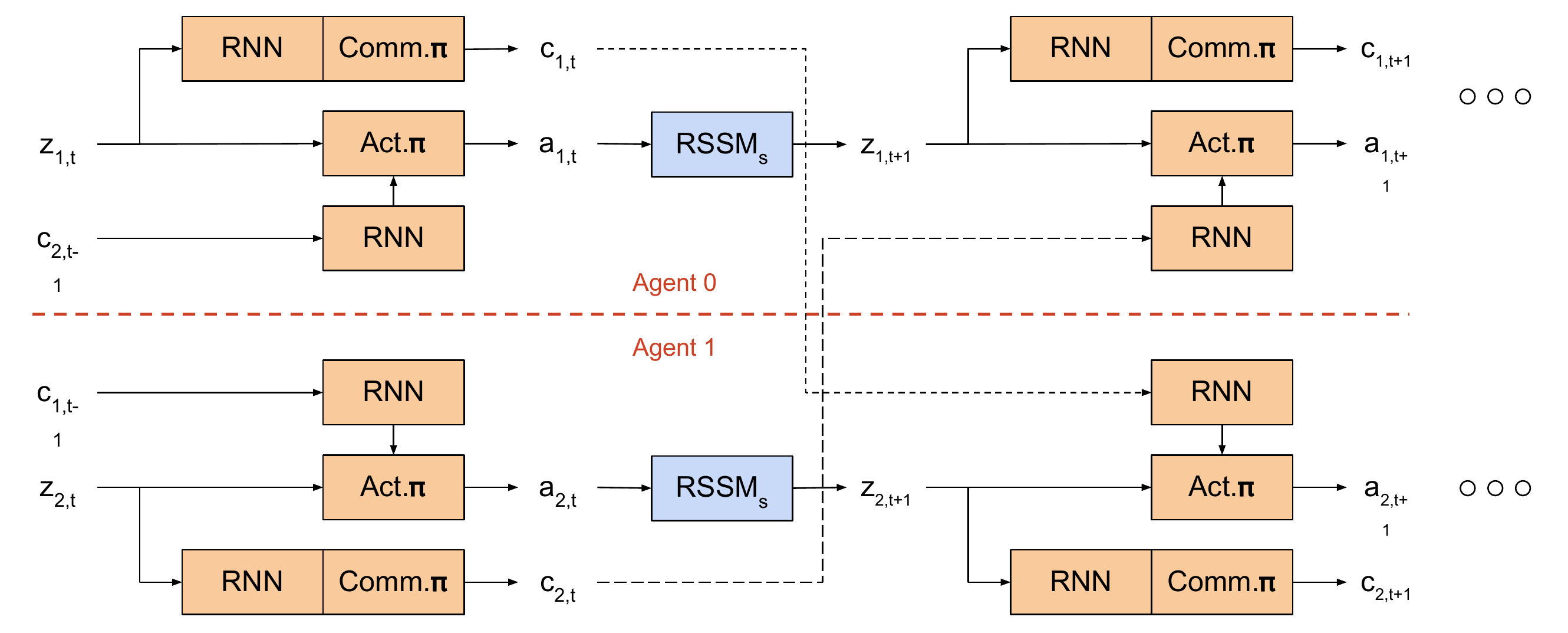}
  \caption{Example of policy and communication learning in imaginary trajectories with two agents. Although communication symbols are not part of world models, they can be learned during imagination rollouts. Orange blocks correspond to components learned through policy learning, where RNNs are included to encode and decode symbols to enable compositionality.}
  \label{fig:imagination-communication}
\end{figure}

Although the RSSMs do not include communication, thanks to the properties of shared imagination, it is possible to learn communications in the imaginary space. This learning occurs in parallel to the normal policy learning as shown in Figure \ref{fig:imagination-communication}, where communication symbols are not used to unroll the RSSM models but are shared among the agents directly, with additional communication encoder/decoders to allow compositional language emergence.

The pipeline resembles the one present in \cite{Mordatch2018}, where having a differentiable model allows for the ease of emergence of communication and coordination. In our method, all components of the rollout are completely learned and differentiable (including the communications), and it is possible to perform policy updates corresponding to several thousand of trajectories every few real environment updates, as in the original Dreamer method.


\section{Experiments}

In this section, we present experimental results for our proposed method, aimed to evaluate the effects of the different components in task appropriate for the target MA settings and evaluate it with similar state-of-the-art methods.

In this context, referential games, like the ones proposed in \cite{referential-image-reinforce}, \cite{referential-pixel}, \cite{communication-multimodal-multistep-referential}, and \cite{communication-qlearning}, provide environments in which communication is essential to solving the tasks, but the agent's actions have little to no effect on the environment state, and episodes consist of a handful of interactions at the most, limiting the analysis that can be performed. Then, \cite{Mordatch2018} proposes a multi-agent navigation task, where communication is required to share goals and solve the task. Although communication is grounded in physical environments rather than symbolic or abstract representations, the physical observability of the agents is complete, and the required coordination can be solved without physical interactions, limiting the scope of the problem. Finally, \cite{coordinationsoccer} provides a physically grounded soccer task, where several agents must coordinate to solve the problem, in a fashion that has long been used to benchmark MA systems (\cite{hfo, robocup}).

In this work, we use a modified version of the tasks proposed in \cite{coordinationsoccer} to emphasize coordination and communication under partial observability. Communication does not arise unless there is a need for it and the lack of full observability makes communication an essential skill, whereas coordination under partial observability is more representative of generic tasks. For these reasons, instead of using full observability for the agents, we use laser-like sensors, that provide a limited amount of information. The resulting environment consists of physically grounded tasks, where continuous and long-term interactions are required, and the partial observability promotes continuous communication in order to achieve effective coordination and solve tasks.  

In particular, we focus on two sub-tasks to evaluate different aspects of our proposed method, where we vary the role and capabilities of the agents but keep the rewards proposed in \cite{coordinationsoccer}.

\textbf{Speaker-listener soccer:} In a similar fashion to \cite{referential-image-reinforce}, \cite{referential-pixel}, \cite{communication-multimodal-multistep-referential}, and \cite{communication-qlearning}, we propose a referential-like game grounded in a physical environment. This task is composed of two agents: one has full observability, but can not act physically on the environment and can only use a communication channel, whereas the second agent can physically interact with the environment and solve the task, but lacks all the information to do so and hence must rely on communication symbols from the other agent. This task acts as resembles normal referential tasks and the navigation problem proposed in \cite{Mordatch2018}, where although the task is physically grounded, the asymmetry of the information implies a complete dependency on the communication channel to solve the task, as the physical agent can not break the symmetries of the field and score to the correct goals. 

\textbf{2-player soccer:} In a similar fashion to \cite{coordinationsoccer}, two agents must cooperate in order to score goals, but with the additional difficulty of the partial observability, which promotes relying on the available 2-way communication channel (both agents send and receive each other's communication symbols). However, unlike in referential games and the navigation problem proposed by \cite{Mordatch2018}, communication is not completely required to solve the task, since we add semantic information to the laser sensors, indicating to which category their readings correspond (e.g., the agent can know which side of the field is detecting with its lasers). This produces a task where communication is beneficial but not the objective itself, as successful yet sub-optimal strategies that do not rely on communication exist. We argue that this is a more representative example of generic tasks, where communication can emerge more naturally compared to artificially forced situations.

To provide fair comparisons with our method, we provide two external baselines in all our problem settings. We compare the performance of MA-Dreamer against MADDPG (\cite{maddpg}), and a naive MA version of the Proximal Policy Optimization (PPO) algorithm (\cite{ppo}). The latter resembles our method in that it performs rollout-based policy updates with recent data; but this method does not possess any mechanism to tackle MA difficulties. The provided results correspond to the social episode reward (sum of the reward of all the involved physical agents) of the best model evaluated for 50 episodes. For our proposed method, we train the agents for 5M agent steps but use 30M steps for the baselines to account for the disadvantages of model-free baselines.

The first of our experiments evaluates the feasibility of the shared imagination method, where policy optimization is performed on a latent space estimated using a different world model than the one used later during evaluation and data collection. To identify the effect of shared imagination without any other additional components involved, we make use of a modified version of the speaker-listener task soccer, where both agents have full observability and no communication occurs. 
Whereas the physical agent learns its own local model and policy, an additional virtual agent learns the global world model, which is used for the shared imagination process. In contrast, the baselines for MADDPG and MA-PPO correspond to their single-agent counterparts due to only being one physical agent. The results presented in Table \ref{tab:referential-global} show that although there is a performance decrease when using shared imagination due to an extra layer of indirection in the imagined MDP used during policy learning, such decrease is minimal, and the corresponding reward is higher than the provided baselines.

\begin{table}[h!]
  \caption{Speaker-listener soccer task with full observability (from the baselines' perspective this task consists of a single-agent 1-player soccer game)}
  \label{tab:referential-global}
  \centering
  \begin{tabular}{llll}
    \toprule
    Algorithm & Shared imagination     & Communications     & Max. avg. reward \\
    \midrule
    MA-PPO & NA & Disabled  & $53.93$     \\
    MADDPG & NA & Disabled  & $31.58$     \\
    \textbf{MA-Dreamer} & \textbf{Disabled}     & \textbf{Disabled} & $\mathbf{78.7}$      \\
    MA-Dreamer & Enabled     & Disabled       & $76.3$  \\
    \bottomrule
  \end{tabular}
\end{table}

In the second experiment, we rely again on the speaker-listener soccer task to evaluate the ability of agents to exploit the communication channel in long-term tasks, this time using the default setting with partial observations for the physical player (listener), turning the problem into a task that can not be solved without relying on the communication channel. 

The results presented in Table \ref{tab:referential-po} show that MA-PPO fails to solve the task, reflecting the difficulties of learning communication-based policies with just policy gradients in long-term problems. On the other hand, MADDPG obtains marginally positive rewards, implying limited communication capabilities, with a reduced performance compared the single-agent case. Finally, the proposed MA-Dreamer is able to solve the task only when using shared imagination, reflecting on its importance for MA settings, albeit with a reduced performance compared to the fully observable case, yet still obtaining higher performance than all baselines in the fully observable experiment.

\begin{table}[h!]
  \caption{Speaker-listener soccer task with limited observability}
  \label{tab:referential-po}
  \centering
  \begin{tabular}{llll}
    \toprule
    Algorithm & Shared imagination     & Communications     & Max. avg. reward \\
    \midrule
    MA-PPO & NA & Enabled  & $-2.69$     \\
    MADDPG & NA & Enabled  & $1.53$     \\
    MA-Dreamer & Disabled     & Enabled & $\sim$NaN      \\
    \textbf{MA-Dreamer} & \textbf{Enabled}     & \textbf{Enabled}       & $\mathbf{42.53}$  \\
    \bottomrule
  \end{tabular}
\end{table}

In the final experiment, we make use of the 2-player soccer task to evaluate the capability of agents to communicate and coordinate in cases where the information that must be shared is not straightforward (unlike the speaker-listener task) and that the information available to each agent is symmetric. 

The results presented in Table \ref{tab:2p-po} show that the baselines can not make use of the communication channel, obtaining similar scores in both with and without the communication channels, and are unable to solve the soccer task. On the other hand, the proposed method obtains similar results compared to the baselines in the case that the communication channel is disabled, but when enabled is able to exploit it and obtain positive rewards.

\begin{table}[h!]
  \caption{2-player task with limited observability}
  \label{tab:2p-po}
  \centering
  \begin{tabular}{llll}
    \toprule
    Algorithm & Shared imagination     & Communications     & Max. avg. reward \\
    \midrule
    MA-PPO & NA & Disabled  & $-2.89$     \\
    MA-PPO & NA & Enabled  & $-2.7$     \\
    MADDPG & NA & Disabled  & $\sim$NaN     \\
    MADDPG & NA & Enabled  & $\sim$NaN     \\
    MA-Dreamer & Disabled     & Disabled & $-7.13$      \\
    MA-Dreamer & Enabled     & Disabled       & $-2.84$  \\
    MA-Dreamer & Disabled     & Enabled & $-2.031$      \\
    \textbf{MA-Dreamer} & \textbf{Enabled}     & \textbf{Disabled}       & $\mathbf{11.8}$  \\
    \bottomrule
  \end{tabular}
\end{table}

\section{Conclusions}


In this paper, we presented a novel model-based approach for MARL that addresses common limitations like the use of outdated transitions (limited performance when using communication channels or partial observations) and centralized critics (rigid agent structure) via the use of a series of world models . By using models and the imagination-based rollouts, policy learning is performed in a fully-differentiable manner using transitions from the current MDP, which makes it easier for the agents to attain improved levels of coordination and communication. Results show that in long-term referential-like tasks, the proposed MA-Dreamer outperforms state-of-the-art methods and that in partially observable tasks with agents possessing symmetric levels of information, only our algorithm is able to achieve some level of cooperation to exploit communication channels in order to improve the social performance.

By solving previous limitations and achieving both coordination and communication skills in less constrained tasks and environments, our method opens the door to further studies involving real populations of agents and how coordination can emerge in these populations by performing imaginary rollouts in small subgroups, with the additional research line of how the presence of a model and the induced full differentiability affects the nature of the emerged communication scheme


\clearpage

\bibliographystyle{unsrtnat}
\bibliography{./mendeley}

\newpage
\appendix


\section{Soccer Environment}
To evaluate coordination and communication under partially observable problems in MARL, we utilize the Deep Mind Soccer Environment first presented in \cite{coordinationsoccer}. In what follows, we detail the changes performed over the original environment in order to provide suitable conditions for our problem setting.

The first modified aspect corresponds to the observation space. Although for cases where we require full observability we use the same observation space as the original work (proprioception and ground-truth-like vision system), for the cases where we evaluate our algorithms under partial observability, we employ a laser-like sensor. It consists of a 64-dimensional array of uniform distance readings with a total field-of-view of $60^\circ$ and a maximum distance of 15 meters.

The second aspect to consider is the action space. Although in the original work the agents can jump in addition to steer and accelerate forward, since our objective is to study cooperation and communication, we simplify the task ignoring the jump actions. Additionally, since we study communications in MA, we add an optional communication channel between agents that consists of discrete symbols (we use a total of 4 different symbols in our experiments to promote compositionality).

Finally, the reward given to each agent is independent of the particular task and corresponds to the same one presented by \cite{coordinationsoccer}, that promotes scoring goals, going towards the ball, and moving the ball towards the opponents' goal, where the particular coefficients for each reward term correspond to the final ones obtained in the same work through population training. In addition, we modify the reward to assign positive rewards only when the agents move forward towards its objectives, to account for the partial observations in our tasks, in contrast to the original work.

In the rest of this Section, we detail particular aspects of the tasks addressed within the soccer environment.

\subsection{Referential Soccer Task}
Referential tasks are standard in MA tasks involving communication among agents and are carried out by two agents with predetermined roles: the speaker and the listener. The speaker has access to information that the listener needs to use in order to solve the task, and so successful communication between agents is a requirement for both agents, as they share a common reward. Although very useful to study the emergence of communication, it usually consists of problems where there is no real physical grounding in the observations through time (communication is usually studied in abstract spaces where the actions of the listener do not have real effects on the next observations)

In this task, we provide a new version of referential tasks, where the speaker (agent with no physical representation in the soccer environment) must continuously provide information to a listener (physical soccer agent in the environment) throughout long episodes and a listener that in addition to the communication channel, also has access to direct but limited information about the environment, and its actions have physical consequences over the environment and condition the information that the listener should send over the next time-steps.

In this case, the amount of environment-information available to the listener is insufficient to solve the tasks, as it corresponds to laser-like sensors, that do not provide enough information to break the symmetry of the field and score to the opponents' goal, recovering the successful exploration of the communication channel as a requisite to solve the task. In addition, since the channel is highly constrained in the number of symbols, the listener must also exploit correctly its limited information in order to achieve high scores, and the speaker must monitor what information will be required by the listener depending on the grounded physical state.

\subsection{2P Soccer Task}
In the second task, we deviate from referential tasks that have predetermined roles and asymmetric information and we propose a simple 2-player cooperative soccer game. In this game, agents have access to limited observations and a bi-directional communication channel to exploit.

However, due to the absence of an agent with the information to break the symmetries product of the partial observations, in this task, we add semantic information to the laser readings to allow the agents to know the direction of the opponent's goal. This information is encoded as an additional dimension for each laser reading, in the form of a 2-dimensional discrete vector (representing whether the reading corresponds to which field line or to a physical object such as the ball or another player).

In this task, although it is possible for an agent to ignore completely its teammate and the communication channel and try to play individually, the limited observability present in the environment makes this approach highly inefficient, and promotes coordination between teammates in the exploration of the environment, searching of the ball and scoring. These properties make this task interesting from the communication point of view since unlike classic referential tasks, the agents do not have predetermined information they must convey in order to solve the task and it is not available at all times due to observability, so they must remember information they need to convey through long time horizons and perform it interactively.

\section{Baselines}
In this Section, we detail aspects related to the design and parameters of the baselines with which our proposed method is compared.

\subsection{MA-PPO}
MA-PPO corresponds to a naive MA version of PPO, in which the previous step communication symbols become part of other agent's observations without any additional MA-related mechanisms. Since the critic learned in PPO is only used as a baseline during policy updates and policy learning is performed through a modified version of standard policy gradients, in principle, this naive MA version of PPO has the potential to work on MA tasks.

Experiments are run for 30M agent steps and data is collected through 8 workers, with updates occurring every 128 steps (each update consists of 4 epochs). Additional training parameters are a PPO clipping parameter of $0.1$, an entropy coefficient of $0.01$ for the policies, a value loss coefficient of $0.5$, and a learning rate of $2.5\times10^{-4}$.

The agent's model consists of an encoder per input observation (e.g., global observation, laser-like observation, incoming communication symbols), whose outputs are then concatenated and processed through an LSTM layer to account for partial observability. Then different heads are created from the LSTM layer output representing the critic and the different policies (action and communication). Since PPO does not restrict the type of policies, action policies are implemented directly as Gaussian distributions, whereas communication policies are implemented as categorical distributions. Each observation encoder consists of 3 fully connected layers with 256 units followed by ReLU activations. Then, the LSTM layer outputs 512 units and each head (critic and policies) are modeled by a single fully connected layer.

\subsection{MADDPG}
The second baseline we consider is MADDPG, an MA-oriented RL algorithm. We follow most of the architecture and design choices of the original work presented in \cite{maddpg}, but adequate some of them to best suit the needs of the addressed tasks.

Data is collected for 30M agent steps through 8 workers (in the original publications, a single worker is used, but in our experiments, the use of multiple workers increases the stability of the learning process). Optimization of the network happens every 100 agent steps and each batch consists of 128 episodes, unlike the 1024 of the original implementation, to account for longer episodes and the computational constraints imposed by LSTM layers.

Since the critic and agents have different inputs, they are modeled by completely independent networks, which share similar architectures. First, inputs are encoded by a fully connected layer and are then concatenated to a simple embedding. Then, they are processed by another fully connected layer and an LSTM layer. Finally, each head (action policy, communication policy, and critic) is modeled through two additional fully connected layers.  All intermediate layers have an output size of 128 units.

Finally, since MADDPG involves deterministic policies, additional explorations perturbations are added. In the case of the action policy, Gaussian noise with $0.1$ is added during data collection, whereas for the communication policy, the policy is used as the logits of a Gumbel-Softmax distribution (\cite{gumbel}) to provide a differential approximation of a discrete distribution for the communication symbols. We keep the temperature fixed through the experiments.

\section{MA-Dreamer Details}
In this paper, we presented MA-Dreamer, a multi-agent version of the model-based method presented in \cite{dreamer2}. In this Section, we provide more details regarding the changes applied to the overall architecture, and the hyper-parameters used during training.

\subsection{Architecture}
As presented in Figures \ref{fig:dreamer-wm-agent} and \ref{fig:dreamer-wm-global}, in our method we include several word-models unlike the original work, that nonetheless follow much of the specifications of the original networks.

The most notable differences are the changes from CNN encoders and decoders utilized for the original Atari and visual control tasks, to multi-layer perceptrons, to account for the nature of our tasks. For both encoder and decoder, we use 3 layers of fully connected layers of 400 units and ReLU activations.

The second change applied to the world model architecture corresponds to the stochastic latent code estimation performed by the global world model (presented in Figure \ref{fig:dreamer-wm-global}). This is implemented as additional $n$-heads parallel to the observation and reward reconstruction, where $n$ is the number of physical players. The latent code estimation is learned through the KL distance in the same way as the transition dynamics, although not bi-directionally, and in the same fashion that the encoder/decoder architecture, we use 3 layers of fully connected layers of 400 units and ReLU activations for the latent code estimation heads.

The last architectural change related to the policy networks. Although the RSSM networks of the world model include memory components, the RSSM itself does not model the communication codes, so additional memory components must be learned during policy learning. As shown in Figure \ref{fig:imagination-communication}, we implement this as additional LSTM layers in the policies: one for the action policy and one for the communication policy of each agent.

The action policies encode the communication codes from other agents through a module composed of 3 fully connected layers and an LSTM layer, each with 64 units. This embedding is then concatenated to the code from the world model, which is then used the generate a policy. In the same way, the communication policy also incorporates an LSTM layer to enable compositional communication and is implemented as 3 fully connected layers followed by an LSTM layer and a last layer to generate logits for a categorical distribution.

Unlike MADDPG, we do not have restrictions when it comes to the nature of the policies, and the action policies are Gaussian distributions with learned variance, whereas the communication policies are categorical and can be learned through the  straight-through
gradient (\cite{straighttrough}) instead of relying on the Gumbel Softmax approximation.

\subsection{Hyper-parameters}
For MA-Dreamer, we train the policies and world model from scratch for 5M agent steps, which is less than the baselines due to earlier convergence attributed to the data efficiency of model-based methods. Training steps are performed every 5 agent steps and consist of one batch of 50 sequences of length 50 for the world model. Policy learning uses the same sequences, but for each of these elements, MA-Dreamer applies an imaginary rollout of length 15, culminating in an effective batch size of 37500 for policy learning.

\end{document}